\documentclass[11pt]{article}

\usepackage[margin=2cm]{geometry}
\usepackage[T1]{fontenc}
\usepackage[utf8]{inputenc}
\usepackage{lmodern}
\usepackage{microtype}
\usepackage{setspace}
\usepackage{graphicx}
\usepackage{caption}
\usepackage{placeins}
\usepackage{booktabs}
\usepackage{array}
\usepackage{longtable}
\usepackage{tabularx}
\usepackage{amsmath,amssymb}
\usepackage{tikz}
\usepackage[numbers,sort&compress]{natbib}
\usepackage[hidelinks]{hyperref}
\usepackage{xurl}
\usetikzlibrary{arrows.meta,positioning}
\DeclareCaptionFont{singlespaced}{\setstretch{1}}

\graphicspath{{figures/}{manuscript/figures/}{../manuscript/figures/}}
\title{Terrain signatures in Welsh settlement names}

\author{
Oktay Karakuş$^{1,*}$ and Can Eyüpoğlu$^{1,2}$
}

\date{}

\begin{document}

\maketitle

\begin{center}
\small
$^{1}$Department of Computer Science and Informatics, Cardiff University,
Cardiff CF24 4AG, UK\\
$^{2}$Department of Computer Engineering, Turkish Air Force Academy,
National Defence University, Istanbul 34149, Türkiye\\[0.5em]
$^{*}$Correspondence: karakuso@cardiff.ac.uk
\end{center}

\begin{abstract}
Landscapes are named, but whether names retain measurable environmental information beyond broad geographic structure is rarely tested. We analysed 3,757 Welsh settlements using a frozen, source-audited 24-element lexical framework, preregistered outcome-specific models and geographically structured validation. The central comparison contrasted 101 settlements carrying high-terrain elements (\textit{bryn} or \textit{mynydd}) with 139 carrying low-terrain elements (\textit{cwm} or \textit{pant}). High-terrain names occupied locations 24.4 m higher relative to their 2-km surroundings (95\% CI, 10.8--38.1 m; Holm-adjusted $p$ = 0.00137). The association remained positive across prespecified 1-, 2- and 5-km neighbourhood definitions and was reproduced using an independently produced elevation source (24.1 m; 95\% CI, 10.6--37.6 m). Adding terrain-name polarity to a non-lexical spatial and settlement baseline reduced geographically held-out mean squared error by 4.63\%, 6.22\% and 7.30\% under 10-, 25- and 50-km spatial blocking, respectively, although improvement varied among held-out regions. River-related names provided weaker, directionally consistent evidence, while the preregistered woodland model was non-estimable. Residual spatial structure, unresolved name language and the absence of independent external replication limit interpretation. Selected Welsh settlement-name categories therefore retain measurable information about present-day terrain within Wales, without establishing individual etymology, causal naming, historical environmental memory or transferability to other naming systems.
\end{abstract}

\section{Introduction}

Place names are among the most persistent forms of spatial information embedded
in cultural landscapes. They remain visible on maps, in administrative systems
and in everyday navigation long after the conditions under which they were
coined may have changed. This persistence makes toponyms potentially valuable
to environmental research, but also difficult to interpret. A place name may
refer to terrain, water, vegetation, land use, ownership, memory or later
linguistic and administrative histories. Toponyms are therefore neither direct
environmental measurements nor neutral labels: naming is culturally mediated
and can also encode authority, language and contested histories of landscape
\citep{Williamson2023}. Their relationship to contemporary environmental
conditions must consequently be tested rather than assumed
\citep{Reszegi2020,RiescoChueca2010,Fuchs2015}.

A growing empirical literature nevertheless shows that place-name information
can support landscape research. Plant-name diversity has been related to
environmental and social factors \citep{FagundezIzco2016}, while toponym
diversity has been associated with vegetation naturalness in conservation
settings \citep{Valko2023}. Toponyms have also been used to interpret rural
landscape character and change \citep{Atik2022,Hearn2024}, to map disappearing
historical landscape features \citep{CalvoIglesias2012}, and, together with GIS
and spatial analysis, to examine cultural-landscape structure
\citep{Zhou2025,Guo2025,MitxelenaHoyos2026}. These studies establish
environmental toponymy as an active empirical field. They do not, however,
justify treating a contemporary surface-form match as a complete etymology, a
direct record of past environmental conditions, or a generally transferable
environmental indicator.

Historical and Welsh-focused scholarship makes this distinction particularly
important. Reconstructions of past land use from place names have required
documentary, topographic and environmental evidence in addition to lexical
interpretation \citep{Conedera2007}. Welsh place-name research similarly draws
on historical forms, mapping and local landscape knowledge rather than relying
on modern surface forms alone \citep{Owen2015,Parry2023}. Welsh toponyms have
layered linguistic histories, locally specific meanings and bilingual
trajectories, and interpretation of an individual name ordinarily requires
evidence beyond its contemporary spelling \citep{Owen2015}.

The question addressed here is therefore deliberately narrower than historical
or etymological reconstruction. We ask whether a predeclared,
source-attributed set of Welsh lexical categories retains measurable
information about specified present-day environmental variables at settlement
locations. The lexical layer is treated as a reproducible exposure definition,
not as a claim that the screening procedure resolves morphology, language
status, historical sense or the environmental reference of every individual
name. This distinction allows environmental correspondence to be tested
without requiring the stronger claim that each detected element constitutes a
validated etymological interpretation.

A second challenge is geographic structure. Nearby settlements can share
terrain, hydrology, settlement history and naming traditions. A conventional
association can therefore arise partly because lexical groups are distributed
differently across space, while random train--test splitting can allow
near-neighbouring observations to share geographic context between model
fitting and evaluation. Spatially structured validation is required when the
claim of interest concerns predictive information at locations separated from
those used for model fitting \citep{Roberts2017,Valavi2019}. Spatially
constrained permutation procedures provide a complementary way to construct
restricted null comparisons while preserving specified geographic structure
\citep{Crabot2019}. Neither approach, however, removes all residual spatial
dependence or converts an observational association into a causal estimate.
Residual spatial autocorrelation may reflect unmeasured processes, spatial
scale or model specification, and should therefore be measured and reported
rather than treated as a minor technical issue
\citep{Legendre1993,Lichstein2002,Dormann2007review,Dormann2007effects,Gaspard2019}.

We address these problems using a declared Wales-wide settlement frame, a
frozen rule-based lexical screen and a source-audited registry of 24 Welsh
place-name elements. Before confirmatory environmental contrasts were examined,
three outcome-specific hypotheses were specified: river proximity (H1),
terrain polarity (H2) and woody cover (H3). The analysis combines
preregistered association models, prespecified sensitivity analyses,
spatially restricted randomisation, residual spatial diagnostics and,
following a documented post-registration amendment, geographically held-out
prediction. For the principal terrain hypothesis, robustness was additionally
examined across alternative neighbourhood definitions, repeated-name controls
and an independently produced elevation source.

The contribution is therefore not the observation that place names can be
related to landscape. Rather, it is a reproducible framework for asking whether
a frozen lexical classification contains environmental information beyond a
stated geographic and settlement baseline, while keeping confirmatory,
robustness and post-registration analyses explicitly separated. Applied to
Welsh settlements, this design provides a bounded test of whether selected
terrain-, water- and woodland-related name categories correspond to
contemporary environmental conditions, and whether the strongest association
retains predictive information when evaluation is geographically separated
within Wales.

\section{Results}

\label{sec:results}

\subsection{Terrain-related settlement names distinguish local topographic position}

\label{sec:results-terrain-primary}

\begin{figure}[htbp]
\centering
\includegraphics[width=0.95\linewidth]{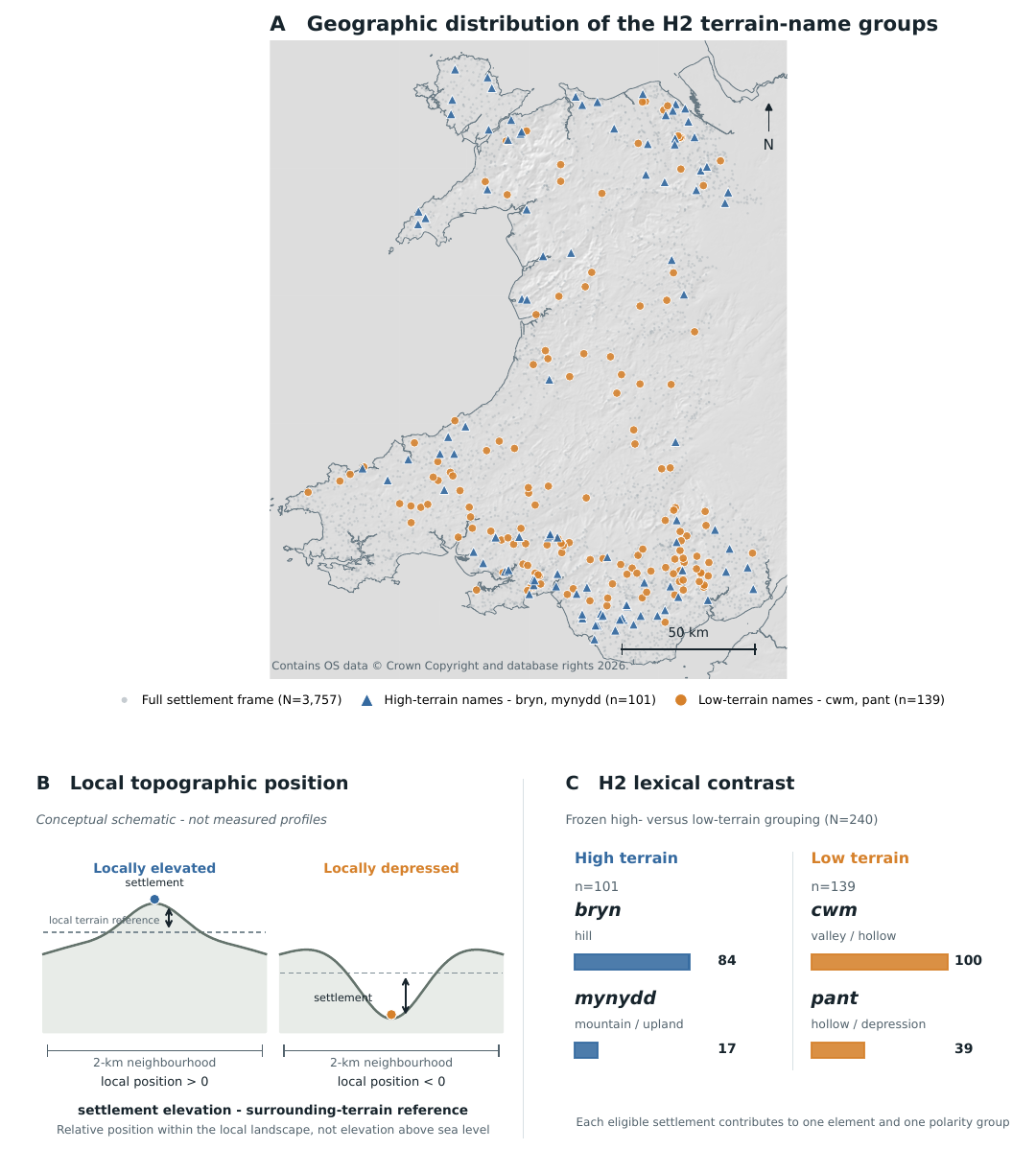}
\captionsetup{font={footnotesize,singlespaced},skip=4pt,labelformat=empty}
\caption{\textbf{Figure 1 | Welsh terrain-name study system and analytical contrast.} \textbf{A,} Geographic distribution of the 240 settlements in the frozen terrain analytical population across Wales. High-terrain names contain \textit{bryn} or \textit{mynydd} (n = 101; blue triangles), whereas low-terrain names contain \textit{cwm} or \textit{pant} (n = 139; orange circles). Small grey points show the complete frozen settlement frame (N = 3,757) for geographic context. \textbf{B,} Conceptual illustration of the primary outcome, local topographic position, defined as settlement elevation minus the surrounding-terrain reference within the primary 2-km neighbourhood. Positive and negative values indicate locally elevated and locally depressed positions, respectively; the schematic does not represent observed terrain profiles or absolute elevation above sea level. \textbf{C,} Terrain lexical contrast and observed element counts: \textit{bryn} (hill; n = 84), \textit{mynydd} (mountain or upland; n = 17), \textit{cwm} (valley or hollow; n = 100), and \textit{pant} (hollow or depression; n = 39). Each eligible record contained one terrain target element and contributed to one polarity group. Contains OS data \textcopyright{} Crown Copyright and database rights 2026.}
\label{fig:study-system}
\end{figure}

Settlement names containing high-terrain elements were associated with systematically higher local topographic positions than names containing low-terrain elements. The preregistered comparison included 240 settlements: 101 containing \textit{bryn} (hill) or \textit{mynydd} (mountain or upland), and 139 containing \textit{cwm} (valley or hollow) or \textit{pant} (hollow or depression). Topographic position was measured as settlement elevation relative to the surrounding terrain within a 2-km neighbourhood, rather than as absolute elevation, allowing the analysis to distinguish whether a settlement occupies a locally elevated or depressed position within its landscape (Fig.~\ref{fig:study-system}).

After adjustment for the prespecified geographic and settlement-level covariates, settlements in the high-terrain group were positioned, on average, \textbf{24.4 m higher relative to their surrounding terrain} than those in the low-terrain group (95\% CI: 10.8--38.1 m; Holm-adjusted \(p = 0.00137\); Fig.~\ref{fig:terrain-contrast-robustness}). The direction of this difference corresponds to the landscape distinction represented by the two lexical groups: settlements whose names contain elements describing elevated terrain occupied higher positions within their local landscapes than settlements whose names contain elements describing valleys, hollows or depressions.

\subsection{The terrain relationship is robust to environmental and analytical choices}

\label{sec:results-terrain-robustness}

\begin{figure}[htbp]
\centering
\includegraphics[width=\linewidth]{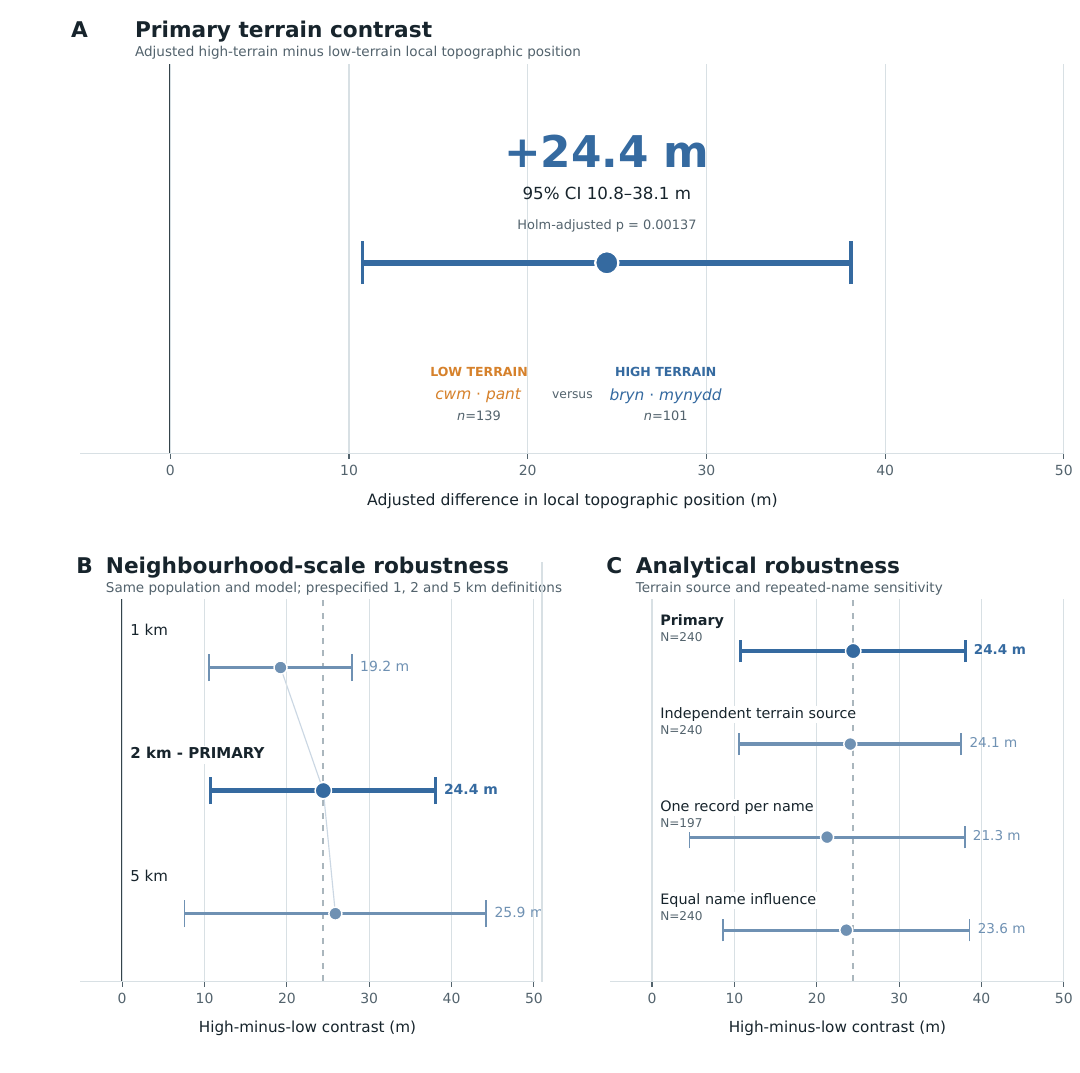}
\captionsetup{font={footnotesize,singlespaced},skip=4pt,labelformat=empty}
\caption{\textbf{Figure 2 | Terrain-name contrast and robustness of local topographic position.} \textbf{A,} Preregistered primary adjusted contrast between settlements with high-terrain lexical elements (\textit{bryn} or \textit{mynydd}) and those with low-terrain elements (\textit{cwm} or \textit{pant}). The high-terrain group occupied positions 24.4 m higher relative to its surrounding terrain than the low-terrain group (95\% CI, 10.8--38.1 m; Holm-adjusted \(p = 0.00137\); N = 240, comprising 101 high-terrain and 139 low-terrain settlements). The estimate is a difference in relative local topographic position, not absolute elevation above sea level. \textbf{B,} Prespecified sensitivity to the neighbourhood used to define local terrain position. Points show adjusted high-minus-low contrasts at 1, 2 and 5 km; the 2-km value is the preregistered primary estimate, not an independent replication. The connecting line links three discrete prespecified analyses and does not represent a fitted scale-response relationship. \textbf{C,} Analytical robustness under an independently produced terrain source, restriction to one record per normalised settlement name and inverse name-group-size weighting. Copernicus GLO-90 is a digital surface model, whereas primary OS Terrain 50 is a digital terrain model; their agreement supports source robustness but not product interchangeability. Points denote adjusted contrasts and horizontal bars show 95\% confidence intervals. Vertical lines mark the zero null and, in robustness panels, the primary 24.4-m estimate as a descriptive reference. Robustness p-values are unadjusted and are not encoded in the figure. Coordinate deduplication is not plotted because all 240 H2 observations occupied distinct coordinate clusters and the procedure removed no observations.}
\label{fig:terrain-contrast-robustness}
\end{figure}

The estimated terrain contrast was not sensitive to the spatial scale used to define local topographic position. Relative to the primary 2-km estimate of 24.4 m, the high-terrain group remained higher than the low-terrain group when surrounding terrain was characterised within either a narrower 1-km neighbourhood (19.2 m) or a broader 5-km neighbourhood (25.9 m). The observed correspondence therefore did not depend on a single choice of neighbourhood scale (Fig.~\ref{fig:terrain-contrast-robustness}B).

The result was similarly insensitive to the source of elevation data. Reconstructing the primary 2-km terrain measure using an independent Copernicus elevation product produced an estimated difference of 24.1 m, closely matching the 24.4-m primary estimate. The terrain contrast was therefore not specific to the elevation product used in the primary analysis (Fig.~\ref{fig:terrain-contrast-robustness}C).

Several analyses addressing dependence among settlement records produced similarly stable estimates. Restricting the analysis to one record per settlement name yielded a 21.3-m difference, while weighting observations so that repeated names contributed equal total influence yielded 23.6 m. A coordinate-cluster deduplication check removed no H2 observations and therefore reproduced the primary 24.4 m estimate exactly. In addition, a spatially restricted permutation test, which preserved local geographic structure while disrupting the correspondence between terrain-name category and topographic position, provided evidence against the null hypothesis of no association (\(p = 0.002\)) (Table~\ref{tab:terrain-robustness-prediction}).

Across these sensitivity analyses, the estimated difference remained positive and of comparable magnitude to the preregistered primary result, indicating that the observed terrain-name correspondence is robust to alternative definitions of local terrain, an independent elevation source and multiple treatments of record-level dependence.

\subsection{Terrain-name information persists beyond geographic structure}

\label{sec:results-terrain-spatial-cv}

The spatial distribution of terrain-related settlement names creates an important alternative explanation for the observed topographic relationship: high- and low-terrain lexical elements may occur preferentially in different parts of Wales, allowing regional geography itself to account for their association with local terrain. We therefore tested whether terrain-name polarity retained predictive information when models were evaluated on geographically held-out settlements.

\begin{figure}[htbp]
\centering
\includegraphics[width=\linewidth]{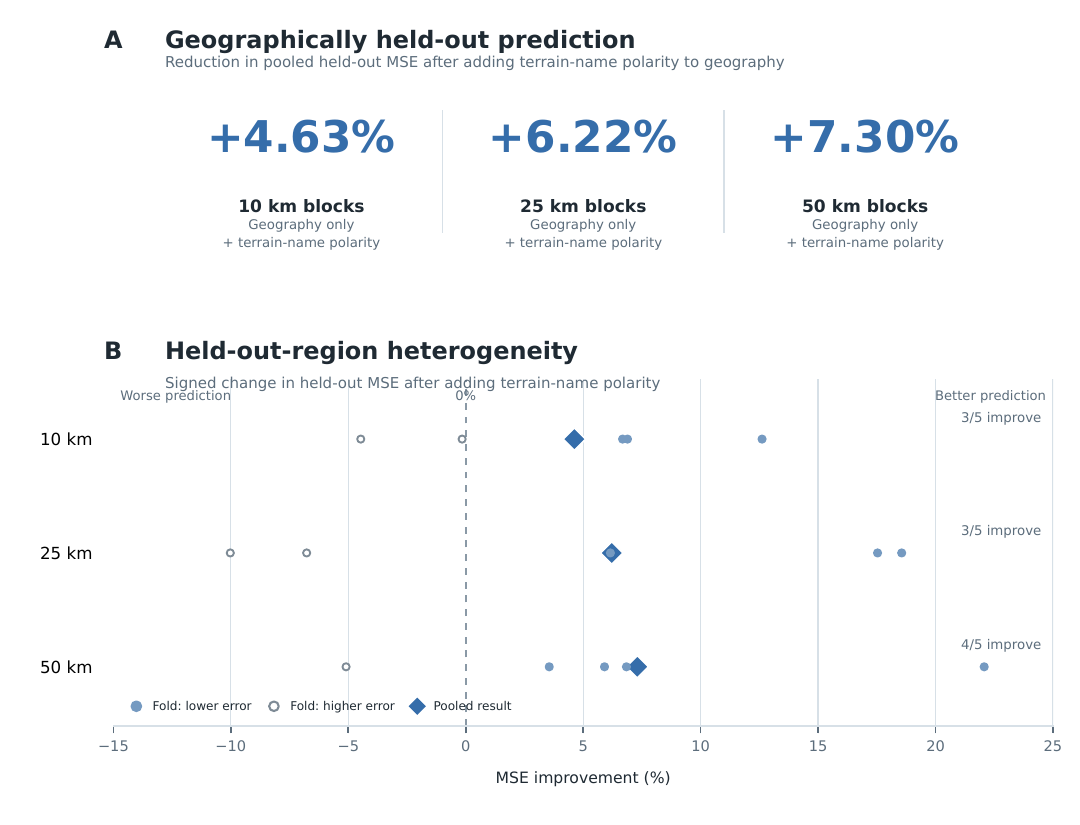}
\captionsetup{font={footnotesize,singlespaced},skip=4pt,labelformat=empty}
\caption{\textbf{Figure 3 | Geographically held-out predictive contribution of Welsh terrain-name polarity.} \textbf{A,} Percentage reduction in pooled held-out mean squared error (MSE) when terrain-name polarity is added to the baseline model. M0 is the non-lexical baseline, comprising settlement-structure covariates and a selected penalised spatial spline; the spatial component is abbreviated as ``Geography only'' in the artwork. M2 adds the H2 high-versus-low terrain-name polarity indicator. Results are shown for three parallel geographically blocked cross-validation evaluations at 10, 25 and 50 km. The displayed progression is not a formal trend across blocking scales. \textbf{B,} Scale-specific held-out fold results. Filled blue circles indicate folds with lower error after adding terrain-name polarity; open grey circles indicate higher error; diamonds show pooled results. Positive values indicate lower held-out MSE after adding terrain-name polarity; negative values indicate higher error. The pooled percentages are calculated from pooled observation-level squared errors and are not averages of the five fold percentages. Three of five folds improved at 10 km, three of five at 25 km and four of five at 50 km. Fold identifiers are scale-specific and are not comparable across scales; fold points illustrate heterogeneity and are not independent replicates for uncertainty estimation. No uncertainty intervals are shown because none were defined in the frozen analysis. This is geographically blocked cross-validation within the Welsh analytical population, not external validation or replication.}
\label{fig:geographically-held-out-prediction}
\end{figure}

Adding the high- versus low-terrain lexical distinction to a model of geographic structure reduced pooled held-out mean squared error by \textbf{4.63\%}, \textbf{6.22\%}, and \textbf{7.30\%} under spatial blocking at 10, 25, and 50 km, respectively. The terrain-name distinction therefore provided additional predictive information across all three spatial blocking scales when evaluation was restricted to geographically separated observations (Fig.~\ref{fig:geographically-held-out-prediction}A).

The improvement was not uniform across individual held-out regions. Adding terrain-name polarity reduced prediction error in three of five folds at both 10 and 25 km and in four of five folds at 50 km, with some folds showing little improvement or higher error. The predictive contribution of terrain-name information should therefore be interpreted as an aggregate out-of-area signal rather than as a geographically uniform effect across Wales (Fig.~\ref{fig:geographically-held-out-prediction}B).

Together with the primary association and sensitivity analyses, these results indicate that the correspondence between terrain-related lexical elements and local topographic position is not fully accounted for by broad geographic structure. Selected Welsh terrain-name elements retain measurable information about local landscape position when prediction is extended to geographically separated settlements.

\begin{table}[!htbp]
\centering
\setcounter{table}{1}
\captionsetup{font={footnotesize,singlespaced},skip=4pt,labelformat=empty}
\caption{\textbf{Table 2 | Robustness and geographically held-out prediction of the terrain-name association.}}
\label{tab:terrain-robustness-prediction}
\footnotesize
\begin{tabularx}{\linewidth}{@{}l c c c c@{}}
\toprule
\multicolumn{5}{@{}l}{\textbf{Panel A | Effect-estimate robustness}}\\
Analysis & $N$ & Adjusted effect (m) & 95\% CI (m) & Raw $p$\\
\midrule
OS Terrain 50, 1 km & 240 & +19.23 & 10.57 to 27.90 & $1.37 \times 10^{-5}$\\
Primary OS Terrain 50, 2 km & 240 & +24.43 & 10.77 to 38.08 & 0.000456\\
OS Terrain 50, 5 km & 240 & +25.89 & 7.56 to 44.22 & 0.00563\\
Copernicus GLO-90, 2 km & 240 & +24.07 & 10.59 to 37.56 & 0.000466\\
One record per normalised name & 197 & +21.25 & 4.54 to 37.97 & 0.0127\\
Equal name-group influence & 240 & +23.59 & 8.62 to 38.56 & 0.00201\\
\midrule
\multicolumn{5}{@{}p{\linewidth}@{}}{\textit{Note.} Coordinate deduplication removed no H2 observations and reproduced the primary estimate exactly. In a complementary restricted spatial randomisation test using 1,999 permutations within spatial/structural strata, the observed +24.43-m coefficient had empirical two-sided $p = 0.002$.}\\
\midrule
\multicolumn{5}{@{}l}{\textbf{Panel B | Geographically held-out prediction}}\\
Blocks & $N$ & M0 MSE & M2 MSE & MSE reduction; improved folds\\
\midrule
10 km & 240 & 1544.72 & 1473.26 & 4.63\%; 3/5\\
25 km & 240 & 1707.35 & 1601.19 & 6.22\%; 3/5\\
50 km & 240 & 1778.14 & 1648.26 & 7.30\%; 4/5\\
\bottomrule
\end{tabularx}
\par\vspace{2pt}\parbox{\linewidth}{\footnotesize\textit{Note.} M0 is the non-lexical baseline comprising settlement-structure covariates and selected penalised spatial-spline terms; M2 additionally includes H2 terrain-name polarity. MSE reduction is calculated from pooled observation-level squared errors and is not the arithmetic mean of fold-level percentage improvements.}
\end{table}

\FloatBarrier

\subsection{Secondary preregistered analyses provide more limited evidence}

\label{sec:results-secondary}

The preregistered river-related analysis provided additional evidence of correspondence between landscape-related lexical information and the local environment. Settlements containing \textit{afon} (river) or \textit{nant} (stream or brook) occurred closer to mapped river networks than settlements without these elements. In the adjusted model, exposure was associated with a multiplicative ratio of 0.649 for $1+\text{river distance}$ (95\% CI: 0.459--0.917; Holm-adjusted
$p=0.0284$). The analysis included 49 exposed settlements and supported the predicted direction of association (Table~\ref{tab:preregistered-tests}).

The third preregistered hypothesis tested whether settlements containing \textit{coed} (wood) or \textit{llwyn} (grove or bush) were associated with greater surrounding woodland cover. The prespecified confirmatory model did not converge and therefore did not yield an interpretable confirmatory estimate. This hypothesis was treated as non-estimable under the preregistered specification rather than as evidence for or against a woodland-name relationship.

\begin{table}[!htbp]
\centering
\setcounter{table}{0}
\captionsetup{font={footnotesize,singlespaced},skip=4pt,labelformat=empty}
\caption{\textbf{Table 1 | Preregistered tests of settlement-name and environmental associations.}}
\label{tab:preregistered-tests}
\footnotesize
\begin{tabularx}{\linewidth}{@{}l>{\raggedright\arraybackslash}X>{\raggedright\arraybackslash}X c >{\raggedright\arraybackslash}X c c >{\raggedright\arraybackslash}X@{}}
\toprule
Hypothesis & Lexical contrast & Outcome & $N$ (groups) & Adjusted effect (95\% CI) & Raw $p$ & Holm $p$ & Result\\
\midrule
H1 & \textit{afon}/\textit{nant} vs other eligible settlements & $\log(1 + \text{river distance})$ & 3,757 (49/3,708) & $-0.433$ ($-0.778$ to $-0.087$) & 0.0142 & 0.0284 & Direction agrees with H1\\
H2 & \textit{bryn}/\textit{mynydd} vs \textit{cwm}/\textit{pant} & 2-km local terrain position & 240 (101/139) & +24.43 m (10.77 to 38.08) & 0.000456 & 0.00137 & Direction agrees with H2\\
H3 & \textit{coed}/\textit{llwyn} vs other eligible settlements & 1-km woody-cover fraction & 2,366 (43/2,323) & Not estimable & --- & 1.000\textsuperscript{\dag} & Model did not converge\\
\bottomrule
\end{tabularx}
\par\vspace{2pt}\parbox{\linewidth}{\footnotesize\textit{Note.} Effects are from the preregistered outcome-specific models with two-sided normalised-name-group cluster-robust standard errors. Holm adjustment was applied across the planned H1--H3 confirmatory family. \textsuperscript{\dag}For H3, the preregistered fractional-logit model did not converge; a value of 1.0 was assigned solely for conservative inclusion in the Holm family and is not an inferential $p$-value. H3 non-estimability should therefore not be interpreted as evidence against the hypothesis.}
\end{table}
\setcounter{table}{2}

\section{Discussion}

The clearest result of this study is that the preregistered terrain-name contrast corresponds to measurable differences in relative position within the local landscape, rather than simply to absolute elevation. Settlements containing \textit{bryn} or \textit{mynydd} occupied positions 24.4 m higher relative to their 2-km surrounding terrain than settlements containing \textit{cwm} or \textit{pant}. The direction of this contrast persisted when local position was defined over 1-, 2- and 5-km neighbourhoods, under an independently produced elevation source, and under analyses reducing the influence of repeated normalised names. These analyses address different potential sources of sensitivity and converge on the same qualitative terrain distinction. The result should therefore be interpreted as evidence that the selected high- and low-terrain lexical categories distinguish present-day local topographic position within the Welsh settlement frame, rather than as evidence that they merely separate settlements occupying different absolute elevations or broad parts of Wales.

The geographically held-out analysis provides a complementary test of that interpretation. Adding H2 terrain-name polarity to a non-lexical baseline containing settlement-structure covariates and a penalised spatial representation reduced pooled held-out mean squared error by 4.63\%, 6.22\% and 7.30\% under 10-, 25- and 50-km spatial blocking, respectively. Importantly, the lexical decomposition shows that this predictive contribution is driven primarily by the preregistered terrain-polarity distinction: the broader non-target lexical indicator added little predictive information at 10 km and worsened prediction at the two broader blocking scales, while adding it to H2 did not produce a stable further gain. This sharpens the interpretation of the spatial validation. The result is not evidence that lexical information in general improves environmental prediction after geography is modelled; rather, the specifically defined terrain polarity retains predictive information not exhausted by the fitted spatial and settlement baseline.

That predictive contribution should nevertheless not be interpreted as geographically uniform. Individual held-out folds varied in both magnitude and direction, with improvement in three of five folds at 10 km, three of five at 25 km and four of five at 50 km. The increasing pooled percentages across blocking scales therefore do not establish a scale-response relationship or show that lexical information becomes more informative with geographic distance. Instead, the blocked analyses provide aggregate evidence that the H2 distinction can contribute out-of-area information within Wales while also demonstrating regional heterogeneity in that contribution. Together with the restricted spatial randomisation result, this makes a stronger case than either conventional model adjustment or random train-test splitting alone, but it does not remove all spatial dependence and does not constitute external validation.

These findings place the contribution within a deliberately narrower interpretation of environmental toponymy. Previous studies have related place-name forms or semantic categories to vegetation, landscape character and historical features \citep{FagundezIzco2016,Valko2023,Atik2022,Hearn2024,CalvoIglesias2012}. The contribution here is neither the observation that place names can correspond to landscape nor an automated reconstruction of individual etymologies. It is a reproducible framework in which lexical exposures are frozen before confirmatory outcome testing, environmental contrasts are defined explicitly, sensitivity to analytical choices can be examined, and geographically structured prediction can test whether a prespecified lexical distinction contains information beyond a stated non-lexical baseline. This distinction is important because correspondence between a modern name category and a modern environmental surface does not, by itself, demonstrate when a name arose, which environmental feature motivated it, or whether the contemporary landscape preserves the conditions under which naming occurred.

The secondary preregistered hypotheses also constrain the scope of the result. River-related \textit{afon} or \textit{nant} settlements occurred closer to mapped river networks in the predicted direction, providing weaker complementary evidence that the approach is not restricted entirely to terrain terminology. However, the exposed group was small and this result does not approach the breadth or robustness of the terrain-polarity evidence. The woodland hypothesis provides a different constraint: the preregistered fractional-logit model for \textit{coed} and \textit{llwyn} did not converge and therefore yielded no interpretable confirmatory estimate. It is consequently neither positive nor negative evidence for a woodland-name association. Taken together, H1--H3 do not support a general proposition that environmental information is recoverable equally across lexical domains. They instead identify terrain polarity as the strongest current empirical result, with river proximity providing secondary evidence and woodland remaining unresolved under the preregistered specification.

Several limitations define what can be inferred next. Residual spatial autocorrelation remains, so neither the adjusted association nor the geographically blocked prediction should be interpreted as causal effects or complete spatial deconfounding. Most analytical settlement names lack an explicit Welsh-language label, and the rule-based lexical screen is not an externally validated morphological or etymological classifier. The comparison between OS Terrain 50 and Copernicus GLO-90 demonstrates robustness to an independently produced terrain source, but the former is a digital terrain model and the latter a digital surface model and their agreement does not establish product interchangeability. More fundamentally, all environmental outcomes are contemporary representations. Demonstrating historical environmental memory would require dated name attestations together with independently reconstructed or dated environmental conditions, while demonstrating linguistic performance or broader geographical generality would require independently constructed reference data and external replication beyond Wales. These are natural extensions of the framework, but they are not conclusions of the present study.

\section{Conclusions}

This study provides a reproducible framework for testing whether a
source-audited lexical layer contains environmental information beyond a stated
geographic and settlement baseline. Within the Welsh settlement frame, the
clearest evidence concerns terrain polarity: settlements containing
\textit{bryn} or \textit{mynydd} occupied systematically higher positions
relative to their local surroundings than settlements containing
\textit{cwm} or \textit{pant}. The contrast persisted across alternative
neighbourhood definitions, an independent elevation source and analyses
reducing repeated-name dependence, and terrain-name polarity also improved
prediction for geographically held-out settlements under three parallel
spatial-blocking designs.

The broader results also define the limits of that conclusion. River-related
names showed weaker evidence in the predicted direction, whereas the
preregistered woodland model was non-estimable. Predictive gains varied among
held-out regions, residual spatial structure remained, and all environmental
outcomes represent contemporary conditions. The results therefore do not
establish causal naming processes, validate individual etymologies, demonstrate
historical environmental memory or show that the framework transfers beyond
Wales.

The contribution is instead methodological and empirical: a frozen,
source-audited lexical classification can be tested against independently
constructed environmental outcomes using preregistered inference, explicit
robustness checks and geographically structured evaluation. In this setting,
selected Welsh terrain-name categories retain measurable information about
present-day local topography that is not exhausted by the fitted spatial and
settlement baseline.

\section{Methods}

\begin{figure}[p]
\centering
\includegraphics[width=0.72\textwidth]{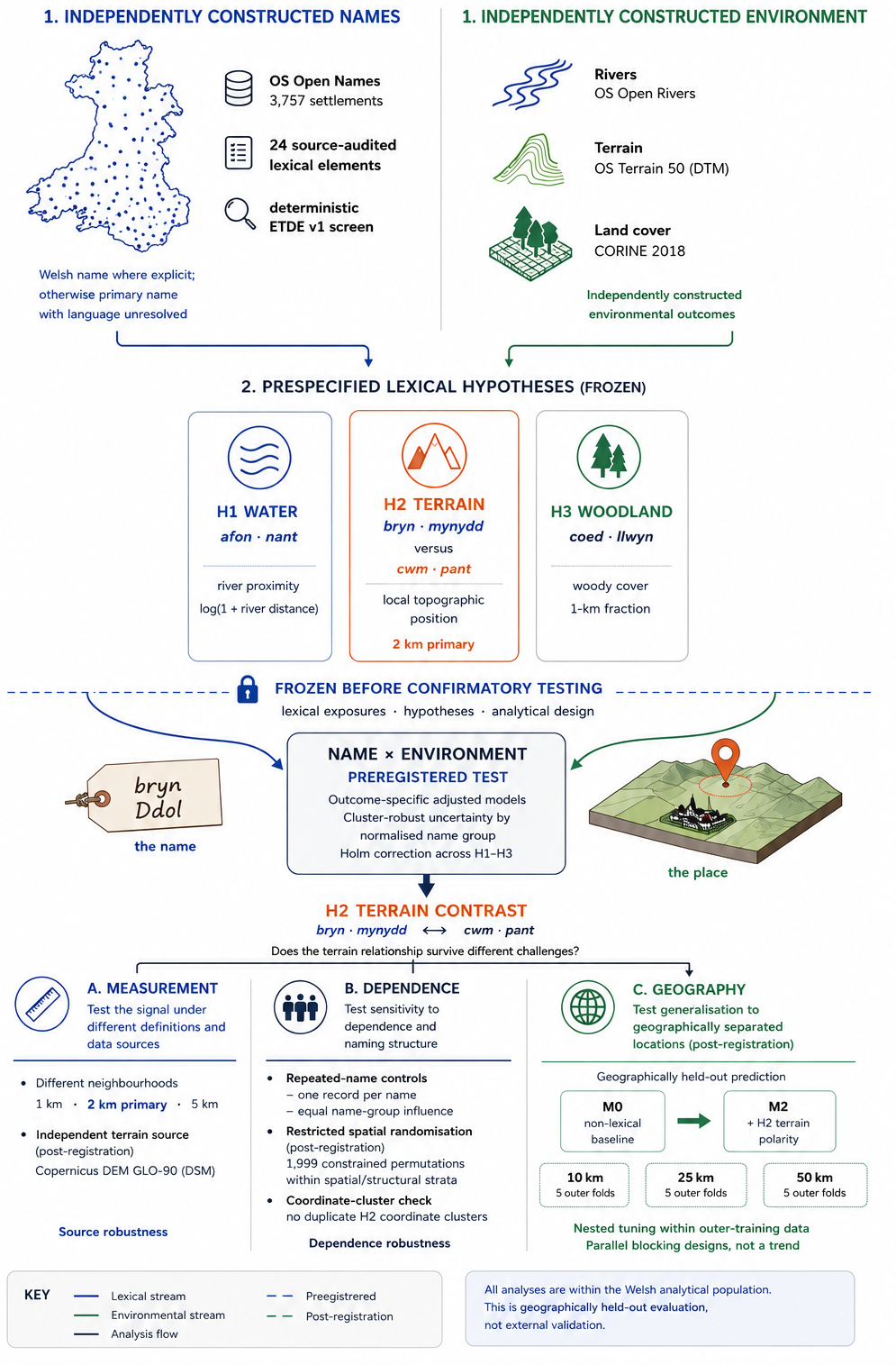}
\captionsetup{font={footnotesize,singlespaced},skip=4pt,labelformat=empty}
\caption{\textbf{Figure 4 | Study design linking Welsh settlement names to environmental evidence.} Lexical exposures and environmental outcomes were constructed independently before confirmatory testing. The frozen lexical system defined preregistered water (H1), terrain (H2) and woodland (H3) hypotheses, which were evaluated using outcome-specific adjusted models with normalised-name-group cluster-robust uncertainty and Holm correction across H1--H3. The principal H2 terrain contrast was subsequently challenged through neighbourhood and independent-source measurement robustness, repeated-name and restricted-randomisation analyses, and geographically held-out prediction. The latter compared the non-lexical baseline M0 with M2, which adds H2 terrain-name polarity, using separate 10-, 25- and 50-km five-fold blocking designs with tuning confined to outer-training observations. Copernicus terrain-source robustness and geographically held-out prediction were post-registration analyses. Spatial validation was conducted within the Welsh analytical population and does not constitute external validation.}
\label{fig:study-design-workflow}
\end{figure}

The complete analytical sequence and the boundary between exposure construction,
environmental outcomes and inference are summarised in Fig.~\ref{fig:study-design-workflow}.

\subsection{Study frame and analytical names}

We analysed 3,757 retained settlement records in Wales from Ordnance Survey Open Names \citep{OSOpenNames2026}. The settlement frame comprised villages, hamlets, towns, suburban areas and other settlements represented as point locations in British National Grid coordinates. We used an explicitly Welsh name when it was labelled as such; otherwise, we used the primary name field and recorded language as unresolved. Repeated settlement names were retained as separate locations and addressed through name-group-clustered inference and prespecified dependence checks.

\subsection{Lexical exposure definition}

We defined the lexical exposures using a fixed list of 24 Welsh place-name elements. For each element, the research record stores its canonical form, a published general gloss, the supporting source, limitations on interpretation and the environmental hypothesis assigned before outcome analysis. The complete list is reported in Supplementary Table S1.

Names were lowercased and stripped of diacritics; hyphens and apostrophes were converted to token boundaries; repeated whitespace was collapsed; and the resulting string was split into tokens. A record was assigned an element when a token began with that element. Exact-token and broader prefix matches were retained separately, allowing the prespecified match-rule sensitivity analyses. This is a deterministic string-screening rule, not a morphological parser, language classifier or semantic disambiguation model. It does not estimate linguistic accuracy, validate individual-name etymology or establish that a matched string carries the recorded general meaning in a particular name.

For software provenance, the frozen screening rule is identified in the repository as ``ETDE v1'' and the machine-readable source registry as ``TETO v1''. The suffix ``v1'' denotes the version fixed before confirmatory analysis; no ETDE v2 was implemented or used. ``TEMPLAR'' is only the repository and preregistration project label, not a separate model, theory or analytical procedure.

Before analysing outcomes, we specified three hypothesis-specific exposures. For H1, we contrasted settlements with an \texttt{afon} or \texttt{nant} match against other river-distance-eligible settlements. For H2, we contrasted settlements with \texttt{bryn} or \texttt{mynydd} matches against those with \texttt{cwm} or \texttt{pant} matches; records containing elements from both contrast groups were excluded. For H3, we contrasted settlements with \texttt{coed} or \texttt{llwyn} matches against other eligible settlements.

\subsection{Environmental outcomes}

The H1 outcome was $\log(1 + \text{river distance})$, where river distance was the planar distance in metres from each settlement to the OS Open Rivers network \citep{OSOpenRivers2025}. The H2 outcome was local terrain position: point elevation minus the mean elevation of raster-cell centres within a circular 2 km buffer, derived from OS Terrain 50 \citep{OSTerrain502025}. The same outcome-blind procedure also calculated 1 km and 5 km terrain-position values for prespecified scale sensitivity. Values were withheld below 95\% valid raster-cell coverage.

The H3 outcome was the 1 km fraction of the buffer covered by CORINE 2018 forest classes or transitional woodland-shrub \citep{CopernicusCLMS2018}. Buffered values required at least 95\% land-cover coverage and the prespecified point-level eligibility flag. No numeric covariates were imputed.

\subsection{Preregistered confirmatory analysis}

The protocol and analysis configuration were externally time-stamped before confirmatory outcome analysis \citep{Karakus2026Preregistration}. H1 and H2 used Gaussian linear models, whereas H3 used a fractional-logit quasi-likelihood model. Each model adjusted for settlement type, analytical-name language status, normalised name-character count, token count, non-target lexical detection and a tensor-product cubic spline of British National Grid easting and northing. The adjustment set was specified to account for settlement characteristics and broad spatial gradients without attempting to remove all local spatial dependence.

Inference used two-sided, normalised-name-group cluster-robust standard errors. H1--H3 formed the planned confirmatory multiplicity family and were adjusted using the Holm procedure. H3 did not yield an estimable confirmatory coefficient because the frozen fractional-logit model failed to converge; for conservative multiplicity accounting it was assigned $p = 1$ within the Holm family, but this value is not an inferential H3 $p$-value. No alternative feasibility model was substituted for the preregistered H3 analysis.

For H1, the exposure coefficient was exponentiated and reported as a ratio in $1 +$ river distance, with values below one indicating shorter adjusted distance among \textit{afon}/\textit{nant} settlements. For H2, the exposure coefficient was reported in metres of local topographic-position difference, defined as high-terrain minus low-terrain settlements. We report analytical sample sizes, exposure-group counts, two-sided 95\% confidence intervals and Holm-adjusted $p$-values for estimable preregistered contrasts.

Prespecified robustness analyses included restricted label permutations, semantic controls, dependence checks, exact-token and initial-token sensitivities, an explicit-language feasibility check, a clean-negative H1 comparison and H2 neighbourhood-scale sensitivity. Residual spatial structure was assessed using Moran's I on model residuals with directed ten-nearest-neighbour weights and 999 permutations. Detailed sensitivity and diagnostic results are reported in the Supplementary Information.

\subsection{Spatially restricted randomisation}

For H2, the prespecified restricted null analysis tested whether the observed high-minus-low terrain-position coefficient was unusual under geographically and structurally constrained reassignment of the exposure labels. We generated 1,999 two-sided permutations by shuffling H2 labels within strata defined by 25-km spatial block, settlement type, analytical-name language status and fixed normalised-name-length bins (1--4, 5--8, 9--12 and $\geq 13$ characters). Environmental outcomes, adjustment covariates, model specification and stratum membership remained fixed. The empirical $p$-value was calculated against the resulting restricted null comparison. This analysis preserves specified local spatial and settlement structure but does not imply complete spatial exchangeability or eliminate residual spatial dependence.

\subsection{Geographically held-out prediction}

The preregistered association models retained residual spatial structure. Following a documented post-registration amendment, we therefore conducted nested, ridge-regularised spatial cross-validation to assess whether lexical information contributed predictive information at geographically separated Welsh settlements. Separate five-fold spatial partitions were constructed for 10-, 25- and 50-km blocking designs. Fold identities are specific to each blocking design and are not interpreted as corresponding geographic regions across scales.

The non-lexical baseline (M0) contained settlement-structure covariates and a spatial spline. Ridge regularisation was applied only to the spatial-spline coefficients; settlement and lexical coefficients remained unpenalised. Penalty selection was performed within the outer-training data, so outer-test observations were not used for model tuning. The corrected H2 comparison (M2) added the high-versus-low terrain-polarity indicator to M0. Held-out performance was evaluated using pooled mean squared error across the outer-test observations. Percentage improvement was calculated as
\[
100 \times \frac{\mathrm{MSE}_{\mathrm{M0}} - \mathrm{MSE}_{\mathrm{M2}}}{\mathrm{MSE}_{\mathrm{M0}}}.
\]
Pooled improvements therefore arise from pooled observation-level squared errors and are not arithmetic means of the five fold-specific percentage improvements.

To determine whether previously reported combined-lexical predictive gains were specifically attributable to H2, a frozen lexical-decomposition analysis additionally compared M1, which adds the non-target lexical indicator to M0, and M3, which contains both H2 polarity and the non-target lexical indicator. M0 and M3 reproduced the corresponding previously frozen spatial-validation outputs, while M2 provides the canonical H2-specific comparison reported in the main text. The decomposition is reported in the Supplementary Information; M3 is not treated as the primary H2 validation result.

The blocked analyses quantify geographically held-out prediction within the Welsh analytical population. They are not an external validation dataset, and the three blocking scales are treated as parallel spatial-validation designs rather than observations defining a continuous distance-response relationship.

\subsection{Terrain-source and dependence robustness}

H2 robustness was evaluated against several changes that preserve the primary terrain-name contrast. The local topographic-position outcome was recalculated using 1- and 5-km neighbourhoods in addition to the preregistered 2-km definition. To assess sensitivity to elevation source, the fixed 2-km procedure was repeated using Copernicus DEM GLO-90, an independently produced digital surface model, whereas the primary OS Terrain 50 product is a digital terrain model \citep{CopernicusGLO90}. This source comparison was conducted post-registration and tests robustness to an alternative elevation product rather than interchangeability of the two products.

Dependence sensitivities included restriction to one record per normalised settlement name and weighting observations so that repeated names contributed equal total influence. A coordinate-cluster deduplication check was also performed; because all 240 H2 observations occupied distinct coordinate clusters, this procedure removed no observations and reproduced the primary result exactly.

\subsection{Use of generative artificial intelligence}

OpenAI ChatGPT/Codex and GitHub Copilot were used to assist with software
development, analysis-workflow support, and manuscript drafting and editing.
All AI-assisted outputs were reviewed and verified by the authors; numerical
results and bibliographic records were checked against the project evidence and
source records. All scientific decisions, interpretation and conclusions remain
the responsibility of the authors.
\section{Data availability}

Publication-safe source data generated in this study, including the numerical
source data underlying the main figures and tables, are available in the
archived reproducibility release at
\url{https://doi.org/10.5281/zenodo.22079109}. Raw third-party spatial data are
not redistributed. The source datasets, access routes and applicable licensing
conditions are documented in the accompanying repository at
\url{https://github.com/oktaykarakus/templar-wales-reproducibility}.

\section{Code availability}

Custom analysis code and reproducibility materials supporting the reported
results are publicly available at
\url{https://github.com/oktaykarakus/templar-wales-reproducibility}. The frozen scientific
release is version \texttt{v1.0.0} and is permanently archived at
\url{https://doi.org/10.5281/zenodo.22079109}. The archived release includes
the claim-bearing analysis code, frozen aggregate outputs, source-data
materials, release manifest, claim-traceability records and SHA-256 checksums.

\section{Supplementary information}

The accompanying Supplementary Information provides detailed study-frame and
lexical provenance, environmental outcome construction, confirmatory and
spatial-analysis methods, robustness analyses, diagnostic evidence and
geographically held-out validation results. Supplementary Tables S1--S7 provide
the lexical inventory, analytical populations, confirmatory and diagnostic
evidence, H2 sensitivity analyses, restricted-null and residual-spatial
diagnostics, spatial-CV lexical decomposition and complete fold-level results.
Claim-essential estimates remain in the main text; the Supplement provides
audit and reproducibility detail rather than a second narrative.

\section{Author contributions}

O.K. conceived and led the study, developed the methodology and software,
curated the data and bibliography, conducted the formal analyses and
visualisation, and wrote the original manuscript. C.E. contributed to
conceptualisation, methodology, investigation, data curation and validation.
Both authors contributed to interpretation of the results, reviewed and edited
the manuscript, and approved the final version.

\section{Competing interests}

The authors declare no competing interests.

\section{Acknowledgements}

The authors thank the organisations that provide and maintain the environmental
and geospatial datasets used in this study.

\bibliographystyle{naturemag}
\bibliography{references}

\end{document}